%% file: main.tex
\documentclass{article} % For LaTeX2e
\usepackage{iclr2020_conference,times}

\input{math_commands.tex}

\usepackage{hyperref}
\usepackage{url}
\usepackage{graphicx}

\usepackage{listings}
\usepackage{xcolor}
 
\definecolor{codegreen}{rgb}{0,0.6,0}
\definecolor{codegray}{rgb}{0.5,0.5,0.5}
\definecolor{codepurple}{rgb}{0.58,0,0.82}
\definecolor{backcolour}{rgb}{0.95,0.95,0.92}
 
\lstdefinestyle{pystyle}{
    backgroundcolor=\color{backcolour},   
    commentstyle=\color{codegreen},
    keywordstyle=\color{magenta},
    numberstyle=\tiny\color{codegray},
    stringstyle=\color{codepurple},
    basicstyle=\ttfamily\tiny,
    emph={and,break,class,continue,def,yield,del,elif ,else,%
        except,exec,finally,for,from,global,if,import,in,%
        lambda,not,or,pass,print,raise,return,try,while,assert,with,as},
    emphstyle=\color{magenta}\bfseries,
    emph={[2]True,False},
    emphstyle=[2]\color{orange}\bfseries,
    breakatwhitespace=false,         
    breaklines=true,                 
    captionpos=b,                    
    keepspaces=true,                 
    numbers=left,    
    numbersep=2pt,
    showspaces=false,                
    showstringspaces=false,
    showtabs=false,                  
    tabsize=2, 
    otherkeywords = {@}
}
 
\title{Streamlining Tensor and Network Pruning in PyTorch}

\author{Michela Paganini \\
Facebook AI Research\\
Menlo Park, CA, USA \\
\texttt{michela@fb.com} \\
\And
Jessica Forde
\thanks{Work completed as part of a Facebook AI Research summer internship.} \\
Brown University \\
Providence, RI, USA \\
\texttt{jessica\_forde@brown.edu} \\
}

\iclrfinalcopy % Uncomment for camera-ready version, but NOT for submission.
\begin{document}

\maketitle

\begin{abstract}
In order to contrast the explosion in size of state-of-the-art machine learning models that can be attributed to the empirical advantages of over-parametrization, and due to the necessity of deploying fast, sustainable, and private on-device models on resource-constrained devices, the community has focused on techniques such as pruning, quantization, and distillation as central strategies for model compression. Towards the goal of facilitating the adoption of a common interface for neural network pruning in PyTorch, this contribution describes the recent addition of the PyTorch \texttt{torch.nn.utils.prune} module, which provides shared, open-source pruning functionalities to lower the technical implementation barrier to reducing model size and capacity before, during, and/or after training.
We present the module's user interface, elucidate implementation details, illustrate example usage, and suggest ways to extend the contributed functionalities to new pruning methods.
\end{abstract}

\section{Introduction}
State-of-the-art deep learning techniques rely on over-parametrized models that are hard to deploy. On the contrary, biological neural networks are known to use efficient sparse connectivity~\citep{Sporns:2007, bhl103261}. Identifying optimal techniques to compress models by reducing the number of parameters is important in order to reduce memory, battery, and hardware consumption without sacrificing accuracy, to deploy lightweight models on device in mobile, IoT, and AR/VR systems, with an eye towards download bandwidth, data consumption, and heat dissipation, and to guarantee privacy with private on-device computation~\citep{nulla, Yang_2017_CVPR, i2016squeezenet, han2015deep, 7551399, kim2015compression, 8110869}.
Real-time applications that require reduced latency, meteorological models that, similar to personalization models, require frequent retraining to capture the latest trends, as well as applications with targeted deployment for custom ASICs, may also benefit from the sparsification of models for train and inference speed concerns. 
Furthermore, the growth in model size has contributed to making reproducing and building upon state-of-the-art techniques only accessible to few, with severe inequalities at a geographical and socioeconomic level. 
Environmental concerns around the cost and carbon footprint of training large-scale models have been documented in~\cite{strubell2019energy, schwartz2019green, henderson2020climate}.

Pruning provides ways to remove unnecessary structure in neural networks, thus beginning to address some of the concerns above. Different ways of identifying superfluous portions of a model result in different pruning techniques. These may vary along several axes, including: the nature of the entities to prune (connections, nodes, channels, layers, etc.), the choice of proxy for importance of each entity (weight, activation, gradient, etc.), when to compute the chosen quantity, the group of entities to pool for comparison (all units in the same layer, the whole network, etc.), when to prune (during, before, or after training), whether pruned entities are forever pruned or can be reinstated, whether to apply hard (binary masks) or soft pruning, iterative or one-shot pruning, and what to do with the network once it is pruned (finetuning, reinitializing, rewinding, etc.).

With the addition of the \texttt{torch.nn.utils.prune} module\footnote{Available at \hyperref[https://github.com/pytorch/pytorch/blob/master/torch/nn/utils/prune.py]{https://github.com/pytorch/pytorch/blob/master/torch/nn/utils/prune.py}}, PyTorch~\citep{pytorch} users may now scan over various choices of pruning techniques as easily as any other choice of hyper-parameter and building block in their machine learning pipeline. At the same time, this module aims at empowering researchers to contribute new pruning techniques and express them through a common language.

The goal of this contribution is knowledge dissemination, among the relevant community, of available tools that can simplify and power both research and deployment in resource-constrained scenarios.

\section{Implementation}

% BasePruningMethod
\texttt{BasePruningMethod} is an abstract base class that provides a skeleton for all pruning techniques and implements shared functionalities. It enables the creation of new pruning techniques by requiring the overriding of methods such as \texttt{compute\_mask}. All pruning methods in Sec.~\ref{sec:pruning_methods} are derived classes that inherit from it.

\begin{figure}
    \centering
    \includegraphics[width=0.5\textwidth]{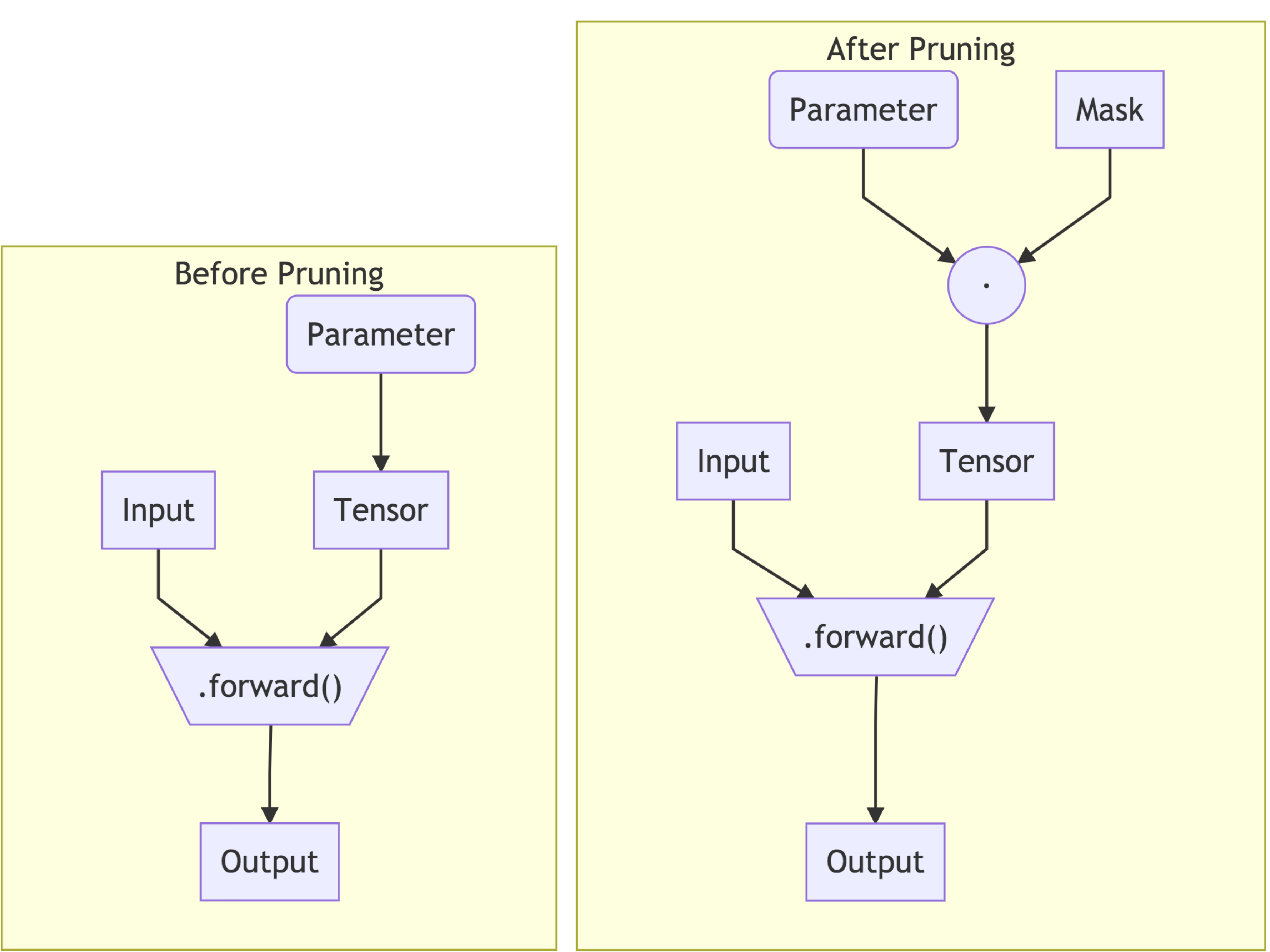}
    \caption{Reparametrization of a tensor in terms of its unpruned version and the computed mask, and its usage in the forward pass.}
    \label{fig:reparameterization}
\end{figure}

The core logic for the application of pruning to a tensor within a module is contained in the the class method \texttt{apply}. Specifically, pruning acts by removing the specified parameter from the parameters list and replacing it with a new parameter whose name equals the original one with the string \texttt{"\_orig"} appended to it. This stores the unpruned version of the tensor. The pruning mask generated by the pruning technique is saved as a module buffer whose name equals the original parameter's name with the string \texttt{"\_mask"} appended to it. Once the reparametrization is in place, an attribute with the original tensor's name, needed by the \texttt{forward} method, is created as a multiplication of the original tensor and the mask by the \texttt{apply\_mask} method. %, which, in turn, is executed as part of the \texttt{apply} class method.
Finally, the pruning method is attached to the module via a \texttt{forward\_pre\_hook} to ensure that the multiplication op (computed on the fly each time upon calling an instance of the pruning class via \texttt{\_\_call\_\_}) is registered into the forward and backward graphs whenever the module is used to compute an output given an input (Fig.~\ref{fig:reparameterization}). The function \texttt{torch.nn.utils.prune.is\_pruned} returns information about the presence of any pruning hook associated with a module. All relevant tensors, including the mask buffers and the original parameters used to compute the pruned tensors are stored in the model's \texttt{state\_dict} and can therefore be easily serialized and saved.

In case the reparametrization and hook creation fail, the related exception is raised and the module is rolled back to its state prior to the failed pruning attempt, without compromising its usability.

To remove the reparametrization depicted in Fig.~\ref{fig:reparameterization} and make the pruning permanent, the user can call the function \texttt{torch.nn.utils.prune.remove}, which removes buffers, hooks, and additional attributes, and assigns the pruned tensor to the parameter with the original parameter name. Note that pruning itself is not undone or reversed by this operation.

A \texttt{PruningContainer} is used to store the history of pruning calls executed on a module in order to enable iterative pruning, \textit{i.e.} the sequential application of pruning techniques. Each parameter in a module that is pruned more than once has an associated \texttt{PruningContainer}; that container has an instance attribute called \texttt{\_tensor\_name} that identifies which parameter in the module it relates to. Only pruning methods that act on the same tensor can be added to the container. The tuple attribute \texttt{\_pruning\_methods} stores the instances of pruning techniques in the order they are applied.

Each pruning method has a \texttt{PRUNING\_TYPE} that dictates how to combine iterative pruning masks. At the moment, this supports the following types: unstructured, structured, and global. An unstructured pruning technique disregards individual entries in a tensor that have already been pruned; a structured pruning technique disregards a row or column only if all its entries have already been pruned; a global pruning method, such a  \texttt{torch.nn.utils.prune.CustomFromMask} in Sec.~\ref{sec:pruning_methods}, applies the pruning technique to all entries, regardless of whether they have been pruned before. These code paths are defined in the \texttt{\_combine\_masks} inner utility function within the \texttt{PruningContainer}'s \texttt{compute\_mask} method. For most other intents and purposes, a container works like any other derived class of a \texttt{BasePruningMethod}.

The \texttt{torch.nn.utils.prune} module also provides a simple and clean functional interface that allows users to interact with pruning techniques through intuitive function calls on modules' parameters identified by name. See Sec.~\ref{sec:example_usage} for example usage.

\subsection{Available Pruning Methods}
\label{sec:pruning_methods}
The following child classes inherit from the \texttt{BasePruningMethod}:
\begin{itemize}
    \item \texttt{torch.nn.utils.prune.Identity}: utility pruning method that does not prune any units but generates the pruning parametrization with a mask of ones;
    \item \texttt{torch.nn.utils.prune.RandomUnstructured}: prune (currently unpruned) entries in a tensor at random;
    \item \texttt{torch.nn.utils.prune.L1Unstructured}: prune (currently unpruned) entries in a tensor by zeroing out the ones with the lowest absolute magnitude;
    \item \texttt{torch.nn.utils.prune.RandomStructured}: prune entire (currently unpruned) rows or columns in a tensor at random;
    \item \texttt{torch.nn.utils.prune.LnStructured}: prune entire (currently unpruned) rows or columns in a tensor based on their $L_n$-norm (supported values of $n$ correspond to supported values for argument $p$ in \texttt{torch.norm()});
    \item \texttt{torch.nn.utils.prune.CustomFromMask}: prune a tensor using a user-provided mask.
\end{itemize}

Their functional equivalents are:
\begin{itemize}
    \item \texttt{torch.nn.utils.prune.identity}
    \item \texttt{torch.nn.utils.prune.random\_unstructured}
    \item \texttt{torch.nn.utils.prune.l1\_unstructured}
    \item \texttt{torch.nn.utils.prune.random\_structured}
    \item \texttt{torch.nn.utils.prune.ln\_structured}
    \item \texttt{torch.nn.utils.prune.custom\_from\_mask}
\end{itemize}

Global pruning, in which entries are compared across multiple tensors, is enabled through \texttt{torch.nn.utils.prune.global\_unstructured} (see Sec.~\ref{sec:example_usage} for example usage).

\subsection{Extending the Module}
The \texttt{torch.nn.utils.prune} module can be extended to implement custom pruning functions. This requires sub-classing the \texttt{BasePruningMethod} base class, and implementing the \texttt{\_\_init\_\_} and \texttt{compute\_mask} methods, \textit{i.e.}, respectively, the constructor and the instructions to compute the mask for the given tensor according to the logic of the pruning technique. The nature of the pruning technique is specified through the assignment of a \texttt{PRUNING\_TYPE}. If none of the currently supported types fits the new pruning technique, the user will also have to add support for a new \texttt{PRUNING\_TYPE} in the way \texttt{PruningContainer} handles the iterative application of pruning masks.

For example, Listing~\ref{list:extend} demonstrates how to implement a pruning technique that prunes every other currently unpruned entry in a tensor, and how to provide a convenient functional interface for the method in just a few lines of code.
\begin{lstlisting}[label={list:extend}, language={Python}, caption={Simple extension to the \texttt{torch.nn.utils.prune} module that implements a custom pruning function to prune every other unpruned entry in a parameter.}]
import torch.nn.utils.prune as prune

class FooBarPruningMethod(prune.BasePruningMethod):
    """Prune every other entry in a tensor
    """
    PRUNING_TYPE = 'unstructured'

    def compute_mask(self, t, default_mask):
        mask = default_mask.clone()
        mask.view(-1)[::2] = 0 
        return mask

def foobar_unstructured(module, name):
    """Prunes tensor corresponding to parameter called `name` in `module`
    by removing every other entry in the tensors.
    Modifies module in place (and also return the modified module) 
    by:
    1) adding a named buffer called `name+'_mask'` corresponding to the 
    binary mask applied to the parameter `name` by the pruning method.
    The parameter `name` is replaced by its pruned version, while the 
    original (unpruned) parameter is stored in a new parameter named 
    `name+'_orig'`.
    Args:
        module (nn.Module): module containing the tensor to prune
        name (string): parameter name within `module` on which pruning
                will act.
    Returns:
        module (nn.Module): modified (i.e. pruned) version of the input
            module
    
    Examples:
        >>> m = nn.Linear(3, 4)
        >>> foobar_unstructured(m, name='bias')
    """
    FooBarPruningMethod.apply(module, name)
    return module
\end{lstlisting}

\section{Example Usage}
\label{sec:example_usage}
Pruning parameters in a model is as simple as invoking the desired pruning function on each parameter (identified by name) within a given neural network module. In this example, the first convolutional layer in a VGG-11 architecture~\citep{simonyan2014deep} is first pruned by removing 3 individual entries using unstructured magnitude-based pruning, then pruned again by removing the bottom 50\% of remaining channels along the 0$^{\mathrm{th}}$ axis by $L_2$-norm.
\begin{lstlisting}[label={simple}, language={Python}, caption={Simple iterative pruning of a single parameter in a network.}]
import torch
from torch import nn
import torch.nn.functional as F
from torchvision.models import vgg11

device = torch.device("cuda" if torch.cuda.is_available() else "cpu")
model = vgg11().to(device=device)

# Prune the first convolutional layer
prune.l1_unstructured(model.features[0], name="weight", amount=3)

# Iteratively prune by simply calling another pruning function on the same parameter.
prune.ln_structured(model.features[0], name="weight", amount=0.5, n=2, dim=0)
\end{lstlisting}
This can be easily extended to apply pruning to all layers in a network. For instance, Listing~\ref{list:2} shows how to prune all 2D convolutional and linear layers in a network, with different pruning fractions that depend on the layer type.
\begin{lstlisting}[label={list:2}, language={Python}, caption={Automated pruning of all weights in a network belonging to specific layer types.}]
new_model = vgg11().to(device=device)
for name, module in new_model.named_modules():
    # Prune 20% of connections in all 2D-conv layers 
    if isinstance(module, torch.nn.Conv2d):
        prune.l1_unstructured(module, name='weight', amount=0.2)
    # Prune 40% of connections in all linear layers 
    elif isinstance(module, torch.nn.Linear):
        prune.l1_unstructured(module, name='weight', amount=0.4)
\end{lstlisting}
In the examples above, each candidate entity for pruning is compared in magnitude to other candidate entities within a single layer. Listing~\ref{list:global}, instead, provides an example of how to enable the pooling together of all entities (\textit{i.e.} single connections, entire units, or channels) across a network for a global magnitude comparison.
\begin{lstlisting}[label={list:global}, language={Python}, caption={Example of how to prune the bottom 20\% of connections by absolute magnitude across an entire LeNet~\citep{lenet} architecture.}]
class LeNet(nn.Module):
    def __init__(self):
        super(LeNet, self).__init__()
        # 1 input image channel, 6 output channels, 3x3 square conv kernel
        self.conv1 = nn.Conv2d(1, 6, 3)
        self.conv2 = nn.Conv2d(6, 16, 3)
        self.fc1 = nn.Linear(16 * 5 * 5, 120)  # 5x5 image dimension
        self.fc2 = nn.Linear(120, 84)
        self.fc3 = nn.Linear(84, 10)

    def forward(self, x):
        x = F.max_pool2d(F.relu(self.conv1(x)), (2, 2))
        x = F.max_pool2d(F.relu(self.conv2(x)), 2)
        x = x.view(-1, int(x.nelement() / x.shape[0]))
        x = F.relu(self.fc1(x))
        x = F.relu(self.fc2(x))
        x = self.fc3(x)
        return x
        
model = LeNet()

parameters_to_prune = (
    (model.conv1, 'weight'),
    (model.conv2, 'weight'),
    (model.fc1, 'weight'),
    (model.fc2, 'weight'),
    (model.fc3, 'weight'),
)

prune.global_unstructured(
    parameters_to_prune,
    pruning_method=prune.L1Unstructured,
    amount=0.2,
)
\end{lstlisting}

The module also supports the pruning of individual tensors by calling the \texttt{prune} method of any pruning class, as shown in Listing~\ref{list:tensor}.
\begin{lstlisting}[label={list:tensor}, language={Python}, caption={Code to prune 70\% of entries at random in a tensor not associated with any \texttt{torch.nn.Module}.}]
t = torch.rand(2, 5)
p = prune.RandomUnstructured(amount=0.7)
pruned_tensor = p.prune(t)
\end{lstlisting}

\bibliography{iclr2020_conference}
\bibliographystyle{iclr2020_conference}

\appendix
% \section{Appendix}

\end{document}

%% file: math_commands.tex
\usepackage{amsmath,amsfonts,bm}

\def\eqref#1{equation~\ref{#1}}
\def\1{\bm{1}}

\DeclareMathAlphabet{\mathsfit}{\encodingdefault}{\sfdefault}{m}{sl}
\SetMathAlphabet{\mathsfit}{bold}{\encodingdefault}{\sfdefault}{bx}{n}